\documentclass[11pt,a4paper]{article}
\usepackage[hyperref]{eacl2021}
\usepackage{times}
\usepackage{latexsym}

\usepackage{float}
\usepackage{subfig}
\usepackage{array}
\usepackage{multirow}
\usepackage[normalem]{ulem}
\usepackage{graphicx}
\usepackage{wrapfig}
 \usepackage{booktabs}
\usepackage[prependcaption,textsize=tiny]{todonotes}

\usepackage{microtype}

\aclfinalcopy 

\newcommand*{\affaddr}[1]{#1} 
\definecolor{ao}{rgb}{0.0, 0.5, 0.0}

\newcommand*{\affmark}[1][*]{\textsuperscript{#1}}
\newcommand*{\email}[1]{\texttt{#1}}

\title{A Little Pretraining Goes a Long Way: A Case Study on Dependency Parsing Task for Low-resource Morphologically Rich Languages}

\author{Jivnesh Sandhan\affmark[1],  Amrith Krishna\affmark[2],  Ashim Gupta\affmark[3],\\
\textbf{Laxmidhar Behera\affmark[1,4]} \textbf{and Pawan Goyal\affmark[5]} \\

\affaddr{\affmark[1]Dept. of Electrical Engineering, IIT Kanpur,}\\
\affaddr{\affmark[2]Dept. of Computer Science and Technology, University of Cambridge,}\\
\affaddr{\affmark[3]School of Computing, University of Utah,}
\affaddr{\affmark[4]Tata Consultancy Services,}\\
\affaddr{\affmark[5]Dept. of Computer Science and Engineering, IIT Kharagpur}\\

\email{jivnesh@iitk.ac.in, ak2329@cam.ac.uk,}
\email{pawang@cse.iitkgp.ac.in}}

\date{}

\begin{document}
\maketitle
\begin{abstract}
Neural dependency parsing has achieved remarkable performance for many domains and languages. The bottleneck of massive labeled data limits the effectiveness of these approaches for low resource languages. In this work, we focus on dependency parsing for morphological rich languages (MRLs) in a low-resource setting. Although morphological information is essential for the dependency parsing task, the morphological disambiguation and lack of powerful analyzers pose challenges to get this information for MRLs. To address these challenges, we propose simple auxiliary tasks for pretraining.
We perform experiments on 10 MRLs in low-resource settings to measure the efficacy of our proposed pretraining method and observe an average absolute gain of 2 points (UAS) and 3.6 points (LAS).\footnote{Code and data available at: \url{https://github.com/jivnesh/LCM}} 
\end{abstract}

\section{Introduction}
\label{intro}

Dependency parsing has greatly benefited from neural network-based approaches. While these approaches simplify the parsing architecture and eliminate the need for hand-crafted feature engineering \cite{chen-manning-2014-fast,dyer-etal-2015-transition,kiperwasser-goldberg-2016-simple,DBLP:conf/iclr/DozatM17,kulmizev-etal-2019-deep}, their performance has been less exciting for several morphologically rich languages (MRLs) and low-resource languages \cite{more-etal-2019-joint,seeker-cetinoglu-2015-graph}. In fact, the need for large labeled treebanks for such systems has adversely affected the development of parsing solutions for low-resource languages~\cite{vania-etal-2019-systematic}. 
\newcite{zeman-etal-2018-conll} observe that data-driven parsing on 9 low resource treebanks resulted not only in low scores but those outputs ``are hardly useful for downstream applications".

Several approaches have been suggested for improving the parsing performance of low-resource languages. This includes data augmentation strategies,  cross-lingual transfer \cite{vania-etal-2019-systematic} and using unlabelled data with semi-supervised learning \cite{clark-etal-2018-semi} and self-training \cite{rotman2019deep}. Further, incorporating morphological knowledge substantially improves the parsing performance for MRLs, including low-resource languages ~\cite{vania-etal-2018-character,dehouck-denis-2018-framework}. This aligns well with the linguistic intuition of the role of morphological markers, especially that of case markers, in deciding the syntactic roles for the words involved \cite{wunderlich2001interaction,sigurdhsson2003case,kittila2011introduction}. However, obtaining the morphological tags for input sentences during run time is a challenge in itself for MRLs \cite{more-etal-2019-joint} and use of predicted tags from taggers, if available, often hampers the performance of these parsers.  In this work, we primarily focus on one such morphologically-rich low-resource language, Sanskrit. 

\begin{figure*}[!tbh]
\centering
\subfloat[\label{fig:tagging_schem}]{\includegraphics[width=0.5\linewidth]{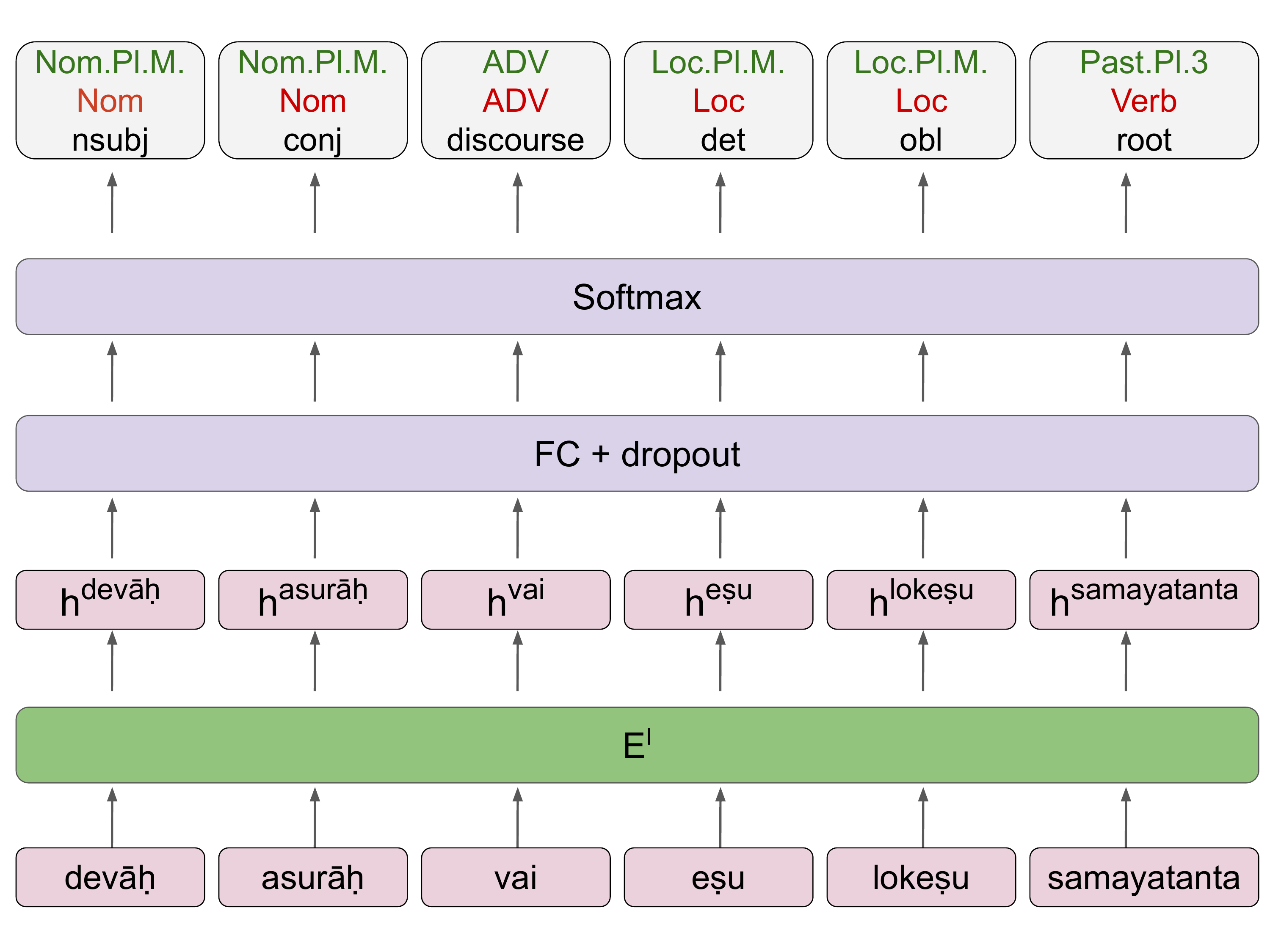}}
\subfloat[\label{fig:gating}]{\includegraphics[width=0.5\linewidth]{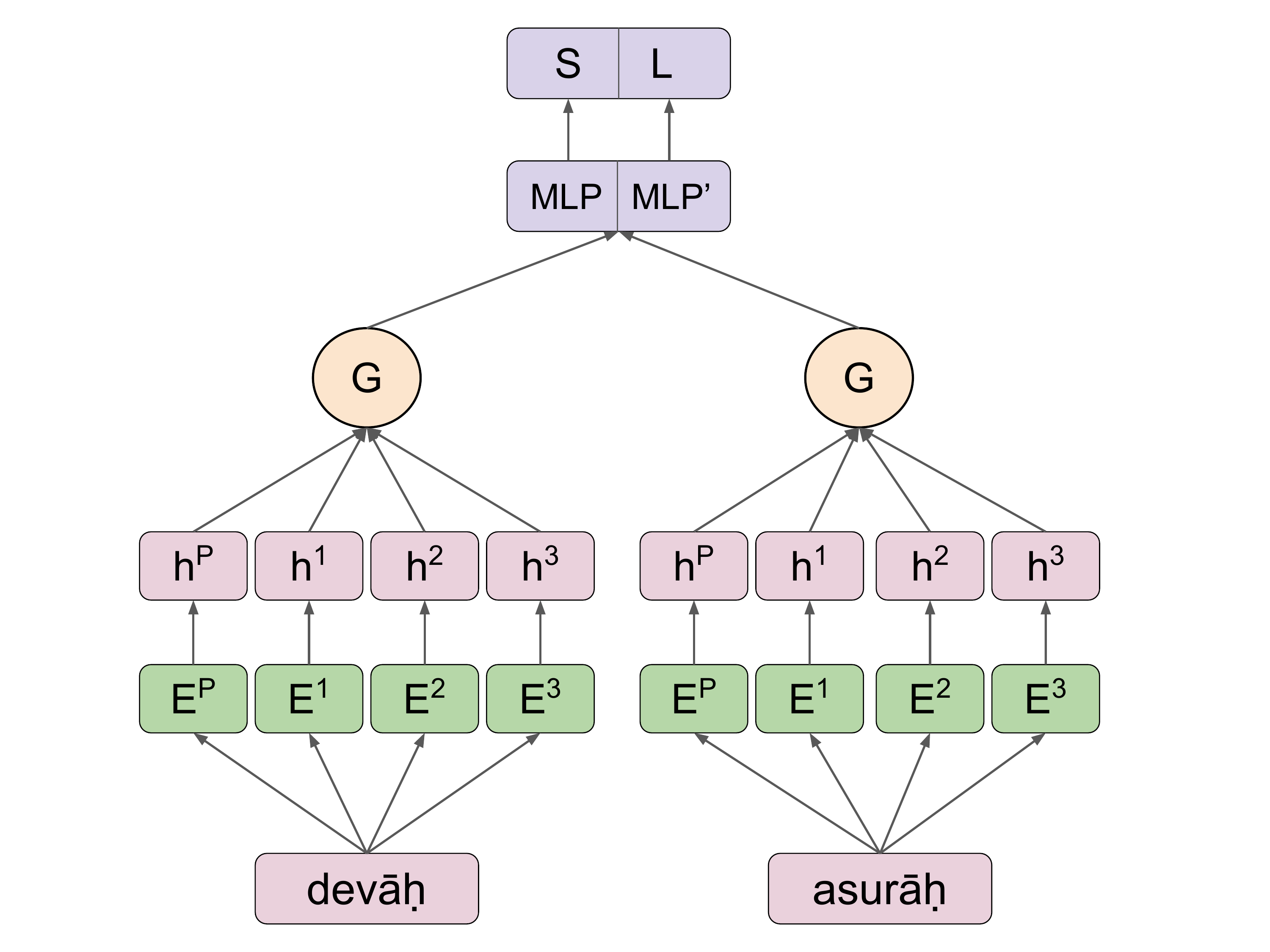}}
\caption{Illustration of proposed architecture for a Sanskrit sequence. English translation: ``Demigods and demons had tried with equal effort for these planets". (a) Pretraining step: For an input word sequence, tagger predicts labels as per three proposed auxiliary tasks, namely, Morphological Tag (green), Case Tag (red) and Label Tag (black). (b) Parser with gating: $E^{(P)}$ is encoder of a neural parser like~\newcite{DBLP:conf/iclr/DozatM17} and $E^{(1)-(3)}$ are the encoders pre-trained with proposed auxiliary tasks. 
Gating mechanism combines representations of all the encoders which,
for each word pair, is passed to two MLPs to predict the probability of arc score (S) and label (L).
}
\label{fig:model}
\end{figure*}

We propose a simple pretraining approach, where we incorporate encoders from simple auxiliary tasks by means of a gating mechanism \cite{sato-etal-2017-adversarial}. This approach outperforms multi-task training and transfer learning methods under the same low-resource data conditions ($\sim$500 sentences). The proposed approach when applied to \newcite{dozat2017stanford}, a neural parser, not only obviates the need for providing morphological tags as input at runtime, but also outperforms its original configuration that uses gold morphological tags as input. Further, our method performs close to DCST \cite{rotman2019deep}, a self-training based extension of \newcite{dozat2017stanford}, which uses gold morphological tags as input for training. 

To measure the efficacy of the proposed method, we further perform a series of experiments on 10 MRLs in low-resource settings and show 2 points and 3.6 points average absolute gain (\S~\ref{results}) in terms of UAS and LAS, respectively. Our proposed method also outperforms multilingual BERT \cite[mBERT]{devlin-etal-2019-bert} based multi-task learning model \cite[Udify]{kondratyuk-straka-2019-75} for the languages which are not covered in mBERT (\S~\ref{Comparison with mBERT Pretraining}).

\section{Pretraining approach}
\label{proposed_model}

Our proposed pretraining approach
essentially attempts to combine word representations from encoders trained on multiple sequence level supervised tasks, as auxiliary tasks, with that of the default encoder of the neural dependency parser. While our approach is generic and can be used with any neural parser,  we use   BiAFFINE parser~\cite{DBLP:conf/iclr/DozatM17}, hence forth referred to as BiAFF, in our experiments.This is a graph-based neural parser that makes use of biaffine attention and a biaffine classifier.\footnote{More details can be found in supplemental (\S~\ref{supple_BiaFF}).} 
Figure~\ref{fig:model} illustrates the proposed approach using an example sequence from Sanskrit. Our pipeline-based approach consists of two steps: (1) Pretraining step (2) Integration step. Figure~\ref{fig:tagging_schem} describes the pretraining step with three auxiliary tasks to pretrain the corresponding encoders $E^{(1)-(3)}$. Finally, in the integration step, these pretrained encoders along with the encoder for the BiAFF model $E^{(P)}$ are then combined using a gating mechanism (\ref{fig:gating}) as employed in ~\newcite{sato-etal-2017-adversarial}. \footnote{Our proposed approach is inspired from \newcite{rotman2019deep}.}

All the auxiliary tasks are trained independently as separate models, but using the same architecture and hyperparameter settings which differ only in terms of the output label they use. The models for the pretraining components are trained using BiLSTM encoders, similar to the encoders in \newcite{DBLP:conf/iclr/DozatM17} and then decoded using two fully connected layers, followed by a softmax layer~\cite{huang2015bidirectional}. These sequential tasks involve prediction of the morphological tag \textbf{(MT)}, dependency label (relation) that each word holds with its head (\textbf{LT}) and further we also consider task where the case information of each nominal forms the output label \textbf{(CT)}. Other grammatical categories did not show significant improvements over the case (\S~\ref{Experiments on additional auxiliary}). This aligns well with the linguistic paradigm that the case information plays an important role in deciding the syntactic role that a nominal can be assigned in the sentence. For words with no case-information, we predict their coarse POS tags. Here, the morphological information is automatically leveraged using the
pre-trained encoders, and thus during runtime the morphological tags need not be provided as inputs. It also helps in reducing the gap between UAS and LAS (\S~\ref{results}).

\section{Experiments}
\label{experiments}
\paragraph{Data and Metric:} We use 500, 1,000 and 1,000 sentences from the Sanskrit Treebank Corpus \cite[STBC]{kulkarni2010designing} as the training, dev and test data respectively for all the models. For the proposed auxiliary tasks, all the sequence taggers are trained with additional previously unused 1,000 sentences from STBC along with the training sentences used for the dependency parsing task. For the Label Tag (LT) prediction auxiliary task, we do not use gold dependency information; rather we use predicted tags from BiAFF parser. For the remaining auxiliary tasks, we use gold standard morphological information.

  For all the models, input representation consists of FastText~\cite{grave-etal-2018-learning}\footnote{\url{https://fasttext.cc/docs/en/crawl-vectors.html}} embedding of 300-dimension and convolutional neural network (CNN) based 100-dimensional character embedding~\cite{NIPS2015_5782}. For character level CNN architecture, we use following setting: 100 number of filters with kernel size equal to 3.
We use standard Unlabelled and Labelled Attachment Scores
(UAS, LAS) to measure the parsing performance and use t-test for statistical significance  ~\cite{dror-etal-2018-hitchhikers}.


 For STBC treebank, the original data does not have morphological tag entry, so the Sanskrit Heritage reader~\cite{huet2013design,goyal2016design} is used to obtain all the possible morphological analysis and only those sentences are chosen which do not have any word showing homonymy or syncretism~\cite{amrith21}.
For other MRLs, we restrict to the same training setup as Sanskrit and use 500 annotated sentences as labeled data for training. Additionally, we use 1000 sentences with morphological information as unlabelled data for pretraining sequence taggers.\footnote{The predicted relations on unlabelled data by the model trained with 500 samples are used for Label Tagging task.} We use all the sentences present in original development and test split data for development and test data. For languages where multiple treebanks are available, we chose only one available treebank to avoid domain shift. Note that STBC adopts a tagging scheme based on the grammatical tradition of Sanskrit, specifically based on K\={a}raka \cite{kulkarni-sharma-2019-paninian,kulkarni2010designing}, while the other MRLs including Sanskrit-Vedic use UD.

 \paragraph{Hyper-parameters:} 
 We utilize the BiAFFINE parser (BiAFF) implemented by~\newcite{ma-etal-2018-stack}. We employ the following hyper-parameter setting for pretraining sequence taggers and base parser BiAFF: the batch size of 16,  number of epochs as 100, and a dropout rate of 0.33 with a learning rate equal to 0.002. The hidden representation generated from n-Stacked-LSTM layers of size 1,024 is passed through two fully connected layers of size 128 and 64. Note that LCM and MTL models use 2-Stacked LSTMs. We keep all the remaining parameters the same as that of~\newcite{ma-etal-2018-stack}.
 
 For all TranSeq variants, one BiLSTM layer is added on top of three augmented pretrained layers from an off-the-shelf morphological tagger~\cite{ashim2020evaluate} to learn task-specific features. In TranSeq-FEA, the dimension of the non-linearity layer of the adaptor module is 256, and in TranSeq-UF, after every 20 epochs, one layer is unfrozen from top to down fashion. In TranSeq-DL, the learning rate is decreased from top to down by a factor of 1.2.
 We have used default parameters to train Hierarchical Tagger~\footnote{\url{https://github.com/ashim95/sanskrit-morphological-taggers}} and baseline models.

\paragraph{Models:} All our experiments are performed as augmentations on two off the shelf neural parsers,  BiAFF~\cite{DBLP:conf/iclr/DozatM17}
and Deep Contextualized Self-training (DCST), which integrates self-training with BiAFF~\cite{rotman2019deep}.\footnote{We describe the baseline models in supplemental (\S~\ref{supple_baselines}).} Hence their default configurations become the baseline models \textbf{(Base)}.
We also use a system that simultaneously trains the BiAFF (and DCST) model for dependency parsing along with the sequence level case prediction task in a multi task setting \textbf{(MTL)}. For MTL model, we also experiment with morphological tagging, as an auxiliary task. However, we do not find significant improvement in performance compared to case tagging.  Hence, we consider case tagging as an auxiliary task to avoid sparsity issue due to the monolithic tag scheme for morphological tagging.
As a transfer learning variant (\textbf{TranSeq}), we extract first three layers from a hierarchical multi-task morphological tagger~\cite{ashim2020evaluate}, trained on 50k examples from DCS~\cite{hellwig2010dcs}. Here each layer corresponds to different grammatical categories, namely, number, gender and case. Note that number of randomly initialised encoder layers in BiAFF (and DCST) are now reduced from 3 to 1. We fine-tune these layers with default learning rate and experiment with four different fine-tuning schedules.\footnote{Refer supplemental (\S~\ref{supple_TranSeq}) for variations of TranSeq.} Finally, our proposed configuration (in \S \ref{proposed_model}) is referred to as the \textbf{LCM} model.\footnote{LCM denotes Label, Case and Morph tagging schemes.} We also train a version each of the base models which expects morphological tags as input and is trained with gold morphological tags. During runtime, we report two different settings, one which uses predicted tags as input (\textbf{Predicted MI}) and other that uses gold tag as input (\textbf{Oracle MI}). We obtain the morphological tags from a Neural CRF tagger \cite{yang2018ncrf} trained on our training data. Oracle MI will act as an upper-bound on the reported results.

\subsection{Results}
\label{results}
Table~\ref{table:san_results} presents results for dependency parsing on Sanskrit.
We observe that BiAFF + LCM outperforms all corresponding BiAFF models including Oracle MI. This is indeed a serendipitous outcome as one would expect Oracle MI to be an upper bound owing to its use of gold morphological tags at runtime. The DCST variant of our pretraining approach is also the best among its peers, although the performance of Oracle MI model in this case is indeed the upper bound.
 \begin{table}[H]
  \begin{small}
\centering
\begin{tabular}{ccccc}
\cmidrule(r){1-5}
 &\multicolumn{2}{c}{BiAFF}
&
\multicolumn{2}{c}{DCST} \\\cmidrule(r){2-3}\cmidrule(l){4-5}
Model &UAS&LAS    & UAS &LAS      \\\cmidrule(r){2-3}\cmidrule(l){4-5}
Base    & 70.67          & 56.85          & 73.23          & 58.64          \\
Predicted MI&69.02&    53.11&    71.15&    51.75\\
\cmidrule(r){2-3}\cmidrule(l){4-5}
MTL     & 70.85          & 57.93          & 73.04          & 59.12          \\
TranSeq  & 71.46 & 60.58 & 74.58 & 62.70\\
LCM    & \textbf{75.91} & \textbf{64.87} & \textbf{75.75} &\textbf{64.28}\\
\cmidrule(r){2-3}\cmidrule(l){4-5}
Oracle MI&\textit{74.08}&    \textit{62.48}&    \textit{76.66}&    \textit{66.35}\\

\hline
\end{tabular}
    \caption{Results on Sanskrit dependency parsing. Oracle MI is an upper bound and is not comparable.}
     \label{table:san_results}
     \end{small}
\end{table}
On the other hand, using predicted morphological tags instead of gold tags at run time degrades results drastically, especially for LAS, possibly due to the cascading effect of incorrect morphological information~\cite{nguyen-verspoor-2018-improved}. 
This shows that morphological information is essential in filling the UAS-LAS gap and substantiates the need for pretraining to incorporate such knowledge even when it is not available at run time. Interestingly, both MTL, and TranSeq, show improvements as compared to the base models, though do not match with that of our pretraining approach. In our experiments, the pretraining approach, even with \textit{a little training data}, clearly outperforms the other approaches.  

 \begin{table}[t]
  \begin{small}
\centering
\begin{tabular}{ccccc}
\hline
Training     & BiAFF       & DCST        & BiAFF+LCM            \\
Size     &UAS/LAS&UAS/LAS&UAS/LAS\\
\hline
100  & 58.0/42.3 & 64.0/44.0 & \textbf{70.4/59.9} \\
500  & 70.7/56.9 & 73.2/58.6 & \textbf{75.9/64.9} \\
750  & 74.0/61.8 & 75.2/62.3 & \textbf{77.3/66.8}  \\
1000 & 74.4/62.9 & 76.0/64.1 & \textbf{77.9/67.3} \\
1250 & 75.6/64.7 & 76.7/65.2 & \textbf{78.5/68.3}\\
\hline
\end{tabular}
    \caption{Performance as a function of training set size.}
     \label{table:ablation_train_size}
     \end{small}
\end{table}

\begin{table*}[bht]
\begin{small}
    \centering
\resizebox{1\textwidth}{!}{%
 \begin{tabular}{ccccccccccc|ccc}
\toprule
 &\multicolumn{2}{c}{eu} &\multicolumn{2}{c}{el} &\multicolumn{2}{c}{sv} &\multicolumn{2}{c}{pl}&\multicolumn{2}{c}{ru}
  &\multicolumn{2}{|c}{avg}
 \\\cmidrule(r){2-3}\cmidrule(l){4-5}\cmidrule(l){6-7}\cmidrule(l){8-9}\cmidrule(l){10-11} \cmidrule(l){12-13}
Model &UAS&LAS    & UAS &LAS   & UAS &LAS & UAS &LAS & UAS &LAS &UAS&LAS    
\\\cmidrule(r){2-3}\cmidrule(l){4-5}\cmidrule(l){6-7}\cmidrule(l){8-9}\cmidrule(l){10-11} \cmidrule(l){12-13}
BiAFF      & 63.18          & 54.52          & 79.64          & 75.01          & 71.73          & 64.83          & 78.33          & 70.83          & 73.98          & 67.42          & 73.37          & 66.52          \\
DCST       & 69.60          & 60.65          & 83.48          & 78.61          & 77.03          & 69.62          & 81.40          & 73.09          & 78.61          & 72.07          & 78.02          & 70.81          \\
\cmidrule(r){2-3}\cmidrule(l){4-5}\cmidrule(l){6-7}\cmidrule(l){8-9}\cmidrule(l){10-11}\cmidrule(l){12-13}
DCST+MTL       & 70.38          & 61.52          & 83.74          & 79.31          & 76.70          & 69.88          & 81.25          & 73.34          & 78.46          & 72.08          & 78.11          & 71.23          \\
DCST+TranSeq   & 70.70          & 62.96          & 84.69          & 80.37          & 77.30          & 70.85          & 82.84          & 75.02          & 78.95          & 73.18          & 78.90          & 72.48          \\
BiAFF+LCM        & \textbf{72.40} & \textbf{65.50} & \textbf{86.56} & \textbf{83.18} & 77.95          & 72.20          & \textbf{84.08} & \textbf{77.65} & 79.97          & 74.47          & 80.20          & 74.60          \\
DCST+LCM       & 72.01          & 65.33          & 85.94          & 82.22          & \textbf{78.72} & \textbf{73.04} & 83.83          & 77.63          & \textbf{80.62} & \textbf{75.26} & \textbf{80.22} & \textbf{74.70} \\
\cmidrule(r){2-3}\cmidrule(l){4-5}\cmidrule(l){6-7}\cmidrule(l){8-9}\cmidrule(l){10-11}\cmidrule(l){12-13}
BiAFF+Oracle MI  & \textit{72.16} & \textit{66.08} & \textit{83.05} & \textit{79.81} & \textit{76.50} & \textit{71.17} & \textit{83.27} & \textit{77.83} & \textit{77.83} & \textit{73.13} & \textit{78.56} & \textit{73.60} \\
DCST+Oracle MI & \textit{77.47} & \textit{71.55} & \textit{85.99} & \textit{82.72} & \textit{80.33} & \textit{75.00} & \textit{86.03} & \textit{80.46} & \textit{82.21} & \textit{77.54} & \textit{82.41} & \textit{77.45}\\
\hline
\\
 &\multicolumn{2}{c}{ar} &\multicolumn{2}{c}{hu} &\multicolumn{2}{c}{fi} &\multicolumn{2}{c}{de}&\multicolumn{2}{c}{cs}
  &\multicolumn{2}{|c}{avg}
 \\\cmidrule(r){2-3}\cmidrule(l){4-5}\cmidrule(l){6-7}\cmidrule(l){8-9}\cmidrule(l){10-11} \cmidrule(l){12-13}
Model &UAS&LAS    & UAS &LAS   & UAS &LAS & UAS &LAS & UAS &LAS &UAS&LAS    
\\\cmidrule(r){2-3}\cmidrule(l){4-5}\cmidrule(l){6-7}\cmidrule(l){8-9}\cmidrule(l){10-11} \cmidrule(l){12-13}
BiAFF      & 76.24          & 68.07          & 70.00          & 62.81          & 60.93          & 50.68          & 67.77          & 59.94          & 65.75          & 57.43          & 70.30          & 62.62          \\
DCST       & 79.05          & 71.18          & 74.62          & 67.00          & 66.04          & 54.76          & 73.22          & 65.18          & 74.15          & 65.52          & 75.61          & 67.70          \\
DCST+Predicted MI & 77.17 & 66.63 & 61.55 & 36.18 & 56.48 & 39.67 & 65.31 & 47.12 & 72.03 & 58.37 & 68.72 & 52.61 \\
\cmidrule(r){2-3}\cmidrule(l){4-5}\cmidrule(l){6-7}\cmidrule(l){8-9}\cmidrule(l){10-11}\cmidrule(l){12-13}
DCST+MTL       & 79.35          & 71.37          & 74.49          & 66.70          & 66.30          & 55.29          & 73.98          & 66.05          & 74.66          & 65.95          & 75.84          & 67.99          \\
DCST+TranSeq-FT   & 79.66          & 72.17          & 75.22          & 68.25          & 67.04          & 56.57          & 74.66          & 67.27          & \textbf{75.15} & 67.02          & 76.40          & 69.11          \\
BiAFF+LCM        & \textbf{79.68} & \textbf{72.55} & \textbf{76.15} & \textbf{69.53} & 69.05          & 59.41          & 75.85          & 68.80          & 74.94          & \textbf{67.58} & 76.91          & 70.13          \\
DCST+LCM       & 79.60          & 72.38          & 75.71          & 68.93          & \textbf{69.15} & \textbf{60.06} & \textbf{76.12} & \textbf{69.20} & 74.81          & 67.54          & \textbf{76.99} & \textbf{70.22} \\
\cmidrule(r){2-3}\cmidrule(l){4-5}\cmidrule(l){6-7}\cmidrule(l){8-9}\cmidrule(l){10-11}\cmidrule(l){12-13}
BiAFF+Oracle MI  & \textit{77.52} & \textit{71.46} & \textit{75.89} & \textit{70.63} & \textit{70.80} & \textit{64.64} & \textit{72.63} & \textit{66.53} & \textit{72.39} & \textit{66.22} & \textit{74.99} & \textit{69.20} \\
DCST+Oracle MI & \textit{80.43} & \textit{74.79} & \textit{78.43} & \textit{73.19} & \textit{75.30} & \textit{68.90} & \textit{77.70} & \textit{71.66} & \textit{78.54} & \textit{72.38} & \textit{79.09} & \textit{73.40}\\
\toprule
\end{tabular}}
    \caption{Evaluation on 10 MRLs. Results of BiAFF+LCM and DCST+LCM are statistically significant compared to strong baseline DCST as per t-test ($p < 0.01$). 
    Last two columns denote the average performance. Models using Oracle MI are not comparable.}
    \label{table:multilingual_results}
    \end{small}
\end{table*}
\paragraph{Ablation:}
We perform further analysis on Sanskrit to study the effect of training set size as well as the impact of various tagging schemes as auxiliary tasks.
First, we evaluate the impact on performance as a function of the training size (Table~\ref{table:ablation_train_size}). 
Noticeably, for training size 100, we observe a 12 (UAS) and 17 (LAS) points increase for BiAFF+LCM over BiAFF, demonstrating the effectiveness of our approach in a very low-resource setting. This improvement is consistent for larger training sizes, though the gain reduces.

\begin{figure}[h]
\centering
\includegraphics[width=0.5\textwidth]{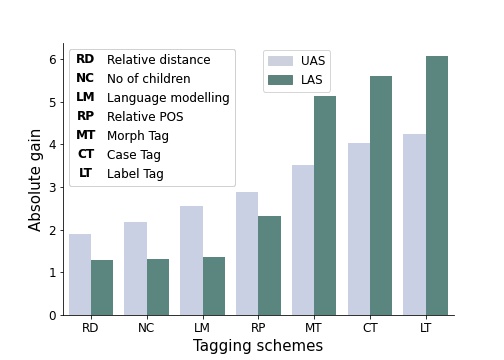}
\caption{Comparison of proposed tagging schemes (MT, CT, LT) with those in DCST (RD, NC, LM, RP).} 
\label{fig:tagging_impact} 
\end{figure}
In Figure~\ref{fig:tagging_impact}, we compare our tagging schemes with those used in self-training of DCST, namely, Relative Distance from root (RD), Number of Children for each word (NC), Language Modeling (LM) objective where task is to predict next word in sentence, and Relative POS (RP) of modifier from root word.  Here, we integrate each pretrained model (corresponding to each tagging scheme) individually on top of the BiAFF baseline using the gating mechanism and report the absolute gain over the BiAFF in terms of UAS and LAS metric.
Interestingly, our proposed tagging schemes, with an improvement of 3-4 points (UAS) and 5-6 points (LAS), outperform those of DCST and help bridge the gap between UAS-LAS.
    \begin{table}[h]
\centering
\begin{tabular}{cccc}
\hline
Auxiliary Task &F-score&Gain\\
\hline
Relative Distance (RD)             & 58.71   & 1.9/1.3 \\
No of children (NC)               & 52.82   & 2.2/1.3 \\
Relative POS  (RP)              & 46.52   & 2.9/2.3 \\
Lang Model (LM)                 & 41.54   & 2.6/1.4 \\
\hline
Coarse POS (CP)                     & 13.02   & 1.6/0.8 \\
Head Word (HW)                   & 40.12   & 1.5/0.4 \\
POS Head Word (PHW)         & 38.98   & 2.0/1.2 \\
Number Tagging (NT)                & 13.33   & 1.9/0.9 \\
Person Tagging (PT)                & 12.27   & 1.6/0.7 \\
Gender Tagging (GT)                & 0.28    & 1.3/0.2 \\
\hline
Morph Tagging (MT)               & 62.84   & 3.5/5.1 \\
Case Tagging (CT)                 & 73.51   & 4.0/5.6 \\
Label Tagging (LT)                & 71.51   & 4.2/6.0\\
\hline
\end{tabular}
    \caption{Comparison of different auxiliary tasks. F-score: Task performance, Gain: Absolute gain (when integrated with BiAFF) in terms of UAS/LAS score compared to BiAFF scores.}
     \label{table:Additional_tag_scheme}
\end{table}
\subsection{Additional auxiliary tasks}
\label{Experiments on additional auxiliary}
  With our proposed pretraining approach, we experiment with using the prediction of different grammatical categories as auxiliary tasks, namely, Number Tagging (NT), Person Tagging (PT), and Gender Tagging (GT). As the results in Table~\ref{table:Additional_tag_scheme} demonstrate, the improvements observed in these cases are much smaller than those for our proposed auxiliary tasks. Similar results are observed when considering other auxiliary tasks (see Table~\ref{table:Additional_tag_scheme}).
We find that combining these auxiliary tasks with our proposed ones did not provide any notable improvements. 
  One possible reason for under performance of these tagging schemes compared to the proposed ones could be that either when the training set is small, sequence taggers are not able to learn discriminative features only from surface form of words  (F-score is less than 40 in all such cases in Table~\ref{table:Additional_tag_scheme}) or the learned features are not helpful for the dependency parsing task.

\subsection{Experiments on other MRLs}
We choose 10 additional MRLs from Universal Dependencies (UD) dataset~\cite{mcdonald2013universal,nivre2016universal}, namely,  
Arabic (ar), Czech (cs), German (de), Basque (eu), Greek (el),    Finnish (fi),    Hungarian (hu), Polish (pl), Russian (ru) and Swedish (sv).\footnote{We choose MRLs that have the explicit morphological information with following grammatical categories: case, number, gender, and tense.}
Then we train them in low-resource setting (500 examples) to investigate the applicability of our approach for these MRLs.

For all MRLs, the trend is similar to what is observed for Sanskrit. While all four models improve over both the baselines, BiAFF+LCM and DCST+LCM consistently turn out to be the best configurations. Note that these models are not directly comparable to Oracle MI models since Oracle MI models use gold morphological tags instead of the predicted ones. 
The performance of BiAFF+LCM and DCST+LCM is also comparable. Across all 11 MRLs, BiAFF+LCM shows the average absolute gain of 2 points (UAS) and 3.6 points (LAS) compared to the strong baseline DCST.

\subsection{Comparison with mBERT Pretraining}
\label{Comparison with mBERT Pretraining}
We compare the proposed method with multilingual BERT \cite[mBERT]{devlin-etal-2019-bert} based multi-task learning  model \cite[Udify]{kondratyuk-straka-2019-75}. This single model trained on 124 UD treebanks covers 75 different languages and produces state of the art results for many of them.
Multilingual BERT leverages large scale pretraining on wikipedia for 104 languages. 
   \begin{table}[H]
  \begin{small}
\centering
\begin{tabular}{cccccc}
\hline
Lang &BiAFF&BiAFF+LCM&Udify\\
\hline
Basque&	63.2/54.5&	72.4/65.5&	\textbf{76.6/69.0}\\
German&	67.7/60.0&	75.8/68.8&	\textbf{83.7/77.5}\\
Hungarian&	70.0/62.8&	76.2/69.5&	\textbf{84.4/76.7}\\
Greek&	69.6/75.0&	86.6/83.2&	\textbf{90.6/87.0}\\
Polish&	78.3.70.8&	84.1/77.7&	\textbf{90.7/85.0}\\
\hline
Sanskrit&	70.7/56.8&\textbf{75.9/64.9}&	69.4/53.2\\
Sanskrit-Vedic&	56.0/42.3& \textbf{61.6/48.0}&	47.4/28.3\\
Wolof&	75.3/67.8&\textbf{78.4/71.3}&		70.9/60.6\\
Gothic&	61.7/53.3&\textbf{69.6/61.4}&		63.4/52.2\\
Coptic&	84.3/80.2&\textbf{86.2/82.7}&		32.7/14.3\\
\hline
\end{tabular}
    \caption{The proposed method outperforms Udify for the languages (down) not covered in mBERT and under performs for the languages (top) which are covered in mBERT.}
     \label{table:mBERT_udify_compare}
     \end{small}
\end{table}
In our experiments, we find that Udify outperforms the proposed method for languages covered during mBERT's  pretraining. Notably, not only the proposed method but also a simple BiAFF parser with randomly initialized embedding outperforms Udify (Table~\ref{table:mBERT_udify_compare}) for languages which not available in mBERT.
Out of 7,000 languages, only a handful of languages can take advantage of mBERT pretraining~\cite{joshi-etal-2020-state} which substantiates the need of our proposed pretraining scheme.

\section{Conclusion}
\label{conclusion}
In this work, we focused on dependency parsing for low-resource MRLs, where getting morphological information itself is a challenge. To address low-resource nature and lack of morphological information, we proposed a simple pretraining method based on sequence labeling that does not require complex architectures or massive labelled or unlabelled data. We show that little supervised pretraining goes a long way compared to transfer learning, multi-task learning, and mBERT pretraining approaches (for the languages not covered in mBERT).
One primary benefit of our approach is that it does not rely on morphological information at run time; instead this information is leveraged using the pretrained encoders. Our experiments across 10 MRLs showed that proposed pretraining provides a significant boost with an average 2 points (UAS) and 3.6 points (LAS) absolute gain compared to DCST.


\section*{Acknowledgements}
We thank Amba Kulkarni for providing Sanskrit dependency treebank data and the anonymous reviewers for their constructive feedback towards improving this work. The work of the first author is supported by the TCS Fellowship under the Project TCS/EE/2011191P. 

\bibliography{eacl2021}

\begin{thebibliography}{45}
\expandafter\ifx\csname natexlab\endcsname\relax\def\natexlab#1{#1}\fi

\bibitem[{Chen and Manning(2014)}]{chen-manning-2014-fast}
Danqi Chen and Christopher Manning. 2014.
\newblock \href {https://doi.org/10.3115/v1/D14-1082} {A fast and accurate
  dependency parser using neural networks}.
\newblock In \emph{Proceedings of the 2014 Conference on Empirical Methods in
  Natural Language Processing ({EMNLP})}, pages 740--750, Doha, Qatar.
  Association for Computational Linguistics.

\bibitem[{Clark et~al.(2018)Clark, Luong, Manning, and
  Le}]{clark-etal-2018-semi}
Kevin Clark, Minh-Thang Luong, Christopher~D. Manning, and Quoc Le. 2018.
\newblock \href {https://doi.org/10.18653/v1/D18-1217} {Semi-supervised
  sequence modeling with cross-view training}.
\newblock In \emph{Proceedings of the 2018 Conference on Empirical Methods in
  Natural Language Processing}, pages 1914--1925, Brussels, Belgium.
  Association for Computational Linguistics.

\bibitem[{Dehouck and Denis(2018)}]{dehouck-denis-2018-framework}
Mathieu Dehouck and Pascal Denis. 2018.
\newblock \href {https://doi.org/10.18653/v1/D18-1312} {A framework for
  understanding the role of morphology in universal dependency parsing}.
\newblock In \emph{Proceedings of the 2018 Conference on Empirical Methods in
  Natural Language Processing}, pages 2864--2870, Brussels, Belgium.
  Association for Computational Linguistics.

\bibitem[{Devlin et~al.(2019)Devlin, Chang, Lee, and
  Toutanova}]{devlin-etal-2019-bert}
Jacob Devlin, Ming-Wei Chang, Kenton Lee, and Kristina Toutanova. 2019.
\newblock \href {https://doi.org/10.18653/v1/N19-1423} {{BERT}: Pre-training of
  deep bidirectional transformers for language understanding}.
\newblock In \emph{Proceedings of the 2019 Conference of the North {A}merican
  Chapter of the Association for Computational Linguistics: Human Language
  Technologies, Volume 1 (Long and Short Papers)}, pages 4171--4186,
  Minneapolis, Minnesota. Association for Computational Linguistics.

\bibitem[{Dozat and Manning(2017)}]{DBLP:conf/iclr/DozatM17}
Timothy Dozat and Christopher~D. Manning. 2017.
\newblock \href {https://openreview.net/forum?id=Hk95PK9le} {Deep biaffine
  attention for neural dependency parsing}.
\newblock In \emph{5th International Conference on Learning Representations,
  {ICLR} 2017, Toulon, France, April 24-26, 2017, Conference Track
  Proceedings}. OpenReview.net.

\bibitem[{Dozat et~al.(2017)Dozat, Qi, and Manning}]{dozat2017stanford}
Timothy Dozat, Peng Qi, and Christopher~D Manning. 2017.
\newblock Stanford’s graph-based neural dependency parser at the conll 2017
  shared task.
\newblock In \emph{Proceedings of the CoNLL 2017 Shared Task: Multilingual
  Parsing from Raw Text to Universal Dependencies}, pages 20--30.

\bibitem[{Dror et~al.(2018)Dror, Baumer, Shlomov, and
  Reichart}]{dror-etal-2018-hitchhikers}
Rotem Dror, Gili Baumer, Segev Shlomov, and Roi Reichart. 2018.
\newblock \href {https://doi.org/10.18653/v1/P18-1128} {The hitchhiker{'}s
  guide to testing statistical significance in natural language processing}.
\newblock In \emph{Proceedings of the 56th Annual Meeting of the Association
  for Computational Linguistics (Volume 1: Long Papers)}, pages 1383--1392,
  Melbourne, Australia. Association for Computational Linguistics.

\bibitem[{Dyer et~al.(2015)Dyer, Ballesteros, Ling, Matthews, and
  Smith}]{dyer-etal-2015-transition}
Chris Dyer, Miguel Ballesteros, Wang Ling, Austin Matthews, and Noah~A. Smith.
  2015.
\newblock \href {https://doi.org/10.3115/v1/P15-1033} {Transition-based
  dependency parsing with stack long short-term memory}.
\newblock In \emph{Proceedings of the 53rd Annual Meeting of the Association
  for Computational Linguistics and the 7th International Joint Conference on
  Natural Language Processing (Volume 1: Long Papers)}, pages 334--343,
  Beijing, China. Association for Computational Linguistics.

\bibitem[{Edmonds(1967)}]{edmonds1967optimum}
Jack Edmonds. 1967.
\newblock Optimum branchings.
\newblock \emph{Journal of Research of the national Bureau of Standards B},
  71(4):233--240.

\bibitem[{Felbo et~al.(2017)Felbo, Mislove, S{\o}gaard, Rahwan, and
  Lehmann}]{felbo-etal-2017-using}
Bjarke Felbo, Alan Mislove, Anders S{\o}gaard, Iyad Rahwan, and Sune Lehmann.
  2017.
\newblock \href {https://doi.org/10.18653/v1/D17-1169} {Using millions of emoji
  occurrences to learn any-domain representations for detecting sentiment,
  emotion and sarcasm}.
\newblock In \emph{Proceedings of the 2017 Conference on Empirical Methods in
  Natural Language Processing}, pages 1615--1625, Copenhagen, Denmark.
  Association for Computational Linguistics.

\bibitem[{French(1999)}]{french1999catastrophic}
Robert~M French. 1999.
\newblock Catastrophic forgetting in connectionist networks.
\newblock \emph{Trends in cognitive sciences}, 3(4):128--135.

\bibitem[{Goyal and Huet(2016)}]{goyal2016design}
Pawan Goyal and G{\'e}rard Huet. 2016.
\newblock Design and analysis of a lean interface for sanskrit corpus
  annotation.
\newblock \emph{Journal of Language Modelling}, 4(2):145--182.

\bibitem[{Grave et~al.(2018)Grave, Bojanowski, Gupta, Joulin, and
  Mikolov}]{grave-etal-2018-learning}
Edouard Grave, Piotr Bojanowski, Prakhar Gupta, Armand Joulin, and Tomas
  Mikolov. 2018.
\newblock \href {https://www.aclweb.org/anthology/L18-1550} {Learning word
  vectors for 157 languages}.
\newblock In \emph{Proceedings of the Eleventh International Conference on
  Language Resources and Evaluation ({LREC} 2018)}, Miyazaki, Japan. European
  Language Resources Association (ELRA).

\bibitem[{Gupta et~al.(2020)Gupta, Krishna, Goyal, and
  Hellwig}]{ashim2020evaluate}
Ashim Gupta, Amrith Krishna, Pawan Goyal, and Oliver Hellwig. 2020.
\newblock Evaluating neural morphological taggers for sanskrit.
\newblock In \emph{Proceedings of the 17th {SIGMORPHON} Workshop on
  Computational Research in Phonetics, Phonology, and Morphology}, Seattle,
  USA. Association for Computational Linguistics.

\bibitem[{Hellwig(2010)}]{hellwig2010dcs}
Oliver Hellwig. 2010.
\newblock Dcs-the digital corpus of sanskrit.
\newblock \emph{Heidelberg (2010-2021). URL
  http://www.sanskrit-linguistics.org/dcs/index.php}.

\bibitem[{Houlsby et~al.(2019)Houlsby, Giurgiu, Jastrzebski, Morrone,
  De~Laroussilhe, Gesmundo, Attariyan, and Gelly}]{pmlr-v97-houlsby19a}
Neil Houlsby, Andrei Giurgiu, Stanislaw Jastrzebski, Bruna Morrone, Quentin
  De~Laroussilhe, Andrea Gesmundo, Mona Attariyan, and Sylvain Gelly. 2019.
\newblock \href {http://proceedings.mlr.press/v97/houlsby19a.html}
  {Parameter-efficient transfer learning for {NLP}}.
\newblock In \emph{Proceedings of the 36th International Conference on Machine
  Learning}, volume~97 of \emph{Proceedings of Machine Learning Research},
  pages 2790--2799, Long Beach, California, USA. PMLR.

\bibitem[{Howard and Ruder(2018)}]{howard-ruder-2018-universal}
Jeremy Howard and Sebastian Ruder. 2018.
\newblock \href {https://doi.org/10.18653/v1/P18-1031} {Universal language
  model fine-tuning for text classification}.
\newblock In \emph{Proceedings of the 56th Annual Meeting of the Association
  for Computational Linguistics (Volume 1: Long Papers)}, pages 328--339,
  Melbourne, Australia. Association for Computational Linguistics.

\bibitem[{Huang et~al.(2015)Huang, Xu, and Yu}]{huang2015bidirectional}
Zhiheng Huang, Wei Xu, and Kai Yu. 2015.
\newblock Bidirectional lstm-crf models for sequence tagging.
\newblock \emph{arXiv preprint arXiv:1508.01991}.

\bibitem[{Huet and Goyal(2013)}]{huet2013design}
G{\'e}rard Huet and Pawan Goyal. 2013.
\newblock Design of a lean interface for sanskrit corpus annotation.
\newblock \emph{Proceedings of ICON}, pages 177--186.

\bibitem[{Joshi et~al.(2020)Joshi, Santy, Budhiraja, Bali, and
  Choudhury}]{joshi-etal-2020-state}
Pratik Joshi, Sebastin Santy, Amar Budhiraja, Kalika Bali, and Monojit
  Choudhury. 2020.
\newblock \href {https://doi.org/10.18653/v1/2020.acl-main.560} {The state and
  fate of linguistic diversity and inclusion in the {NLP} world}.
\newblock In \emph{Proceedings of the 58th Annual Meeting of the Association
  for Computational Linguistics}, pages 6282--6293, Online. Association for
  Computational Linguistics.

\bibitem[{Kiperwasser and Goldberg(2016)}]{kiperwasser-goldberg-2016-simple}
Eliyahu Kiperwasser and Yoav Goldberg. 2016.
\newblock \href {https://doi.org/10.1162/tacl_a_00101} {Simple and accurate
  dependency parsing using bidirectional {LSTM} feature representations}.
\newblock \emph{Transactions of the Association for Computational Linguistics},
  4:313--327.

\bibitem[{Kittil{\"a} et~al.(2011)Kittil{\"a}, V{\"a}sti, and
  Ylikoski}]{kittila2011introduction}
Seppo Kittil{\"a}, Katja V{\"a}sti, and Jussi Ylikoski. 2011.
\newblock Introduction to case, animacy and semantic roles.
\newblock \emph{Case, animacy and semantic roles}, 99:1--26.

\bibitem[{Kondratyuk and Straka(2019)}]{kondratyuk-straka-2019-75}
Dan Kondratyuk and Milan Straka. 2019.
\newblock \href {https://doi.org/10.18653/v1/D19-1279} {75 languages, 1 model:
  Parsing {U}niversal {D}ependencies universally}.
\newblock In \emph{Proceedings of the 2019 Conference on Empirical Methods in
  Natural Language Processing and the 9th International Joint Conference on
  Natural Language Processing (EMNLP-IJCNLP)}, pages 2779--2795, Hong Kong,
  China. Association for Computational Linguistics.

\bibitem[{Krishna et~al.(2020)Krishna, Santra, Gupta, Satuluri, and
  Goyal}]{amrith21}
Amrith Krishna, Bishal Santra, Ashim Gupta, Pavankumar Satuluri, and Pawan
  Goyal. 2020.
\newblock \href {https://doi.org/10.1162/coli\_a\_00390} {A graph based
  framework for structured prediction tasks in sanskrit}.
\newblock \emph{Computational Linguistics}, 46(4):1--63.

\bibitem[{Kulkarni et~al.(2010)Kulkarni, Pokar, and
  Shukl}]{kulkarni2010designing}
Amba Kulkarni, Sheetal Pokar, and Devanand Shukl. 2010.
\newblock Designing a constraint based parser for sanskrit.
\newblock In \emph{International Sanskrit Computational Linguistics Symposium},
  pages 70--90. Springer.

\bibitem[{Kulkarni and Sharma(2019)}]{kulkarni-sharma-2019-paninian}
Amba Kulkarni and Dipti Sharma. 2019.
\newblock \href {https://doi.org/10.18653/v1/W19-7724} {Paninian
  syntactico-semantic relation labels}.
\newblock In \emph{Proceedings of the Fifth International Conference on
  Dependency Linguistics (Depling, SyntaxFest 2019)}, pages 198--208, Paris,
  France. Association for Computational Linguistics.

\bibitem[{Kulmizev et~al.(2019)Kulmizev, de~Lhoneux, Gontrum, Fano, and
  Nivre}]{kulmizev-etal-2019-deep}
Artur Kulmizev, Miryam de~Lhoneux, Johannes Gontrum, Elena Fano, and Joakim
  Nivre. 2019.
\newblock \href {https://doi.org/10.18653/v1/D19-1277} {Deep contextualized
  word embeddings in transition-based and graph-based dependency parsing - a
  tale of two parsers revisited}.
\newblock In \emph{Proceedings of the 2019 Conference on Empirical Methods in
  Natural Language Processing and the 9th International Joint Conference on
  Natural Language Processing (EMNLP-IJCNLP)}, pages 2755--2768, Hong Kong,
  China. Association for Computational Linguistics.

\bibitem[{Ma et~al.(2018)Ma, Hu, Liu, Peng, Neubig, and
  Hovy}]{ma-etal-2018-stack}
Xuezhe Ma, Zecong Hu, Jingzhou Liu, Nanyun Peng, Graham Neubig, and Eduard
  Hovy. 2018.
\newblock \href {https://doi.org/10.18653/v1/P18-1130} {Stack-pointer networks
  for dependency parsing}.
\newblock In \emph{Proceedings of the 56th Annual Meeting of the Association
  for Computational Linguistics (Volume 1: Long Papers)}, pages 1403--1414,
  Melbourne, Australia. Association for Computational Linguistics.

\bibitem[{McCloskey and Cohen(1989)}]{MCCLOSKEY1989109}
Michael McCloskey and Neal~J. Cohen. 1989.
\newblock \href {https://doi.org/https://doi.org/10.1016/S0079-7421(08)60536-8}
  {Catastrophic interference in connectionist networks: The sequential learning
  problem}.
\newblock In Gordon~H. Bower, editor, \emph{Psychology of Learning and
  Motivation}, volume~24, pages 109 -- 165. Academic Press.

\bibitem[{McDonald et~al.(2013)McDonald, Nivre, Quirmbach-Brundage, Goldberg,
  Das, Ganchev, Hall, Petrov, Zhang, T{\"a}ckstr{\"o}m
  et~al.}]{mcdonald2013universal}
Ryan McDonald, Joakim Nivre, Yvonne Quirmbach-Brundage, Yoav Goldberg, Dipanjan
  Das, Kuzman Ganchev, Keith Hall, Slav Petrov, Hao Zhang, Oscar
  T{\"a}ckstr{\"o}m, et~al. 2013.
\newblock Universal dependency annotation for multilingual parsing.
\newblock In \emph{Proceedings of the 51st Annual Meeting of the Association
  for Computational Linguistics (Volume 2: Short Papers)}, pages 92--97.

\bibitem[{More et~al.(2019)More, Seker, Basmova, and
  Tsarfaty}]{more-etal-2019-joint}
Amir More, Amit Seker, Victoria Basmova, and Reut Tsarfaty. 2019.
\newblock \href {https://doi.org/10.1162/tacl_a_00253} {Joint transition-based
  models for morpho-syntactic parsing: Parsing strategies for {MRL}s and a case
  study from modern {H}ebrew}.
\newblock \emph{Transactions of the Association for Computational Linguistics},
  7:33--48.

\bibitem[{Nguyen and Verspoor(2018)}]{nguyen-verspoor-2018-improved}
Dat~Quoc Nguyen and Karin Verspoor. 2018.
\newblock \href {https://doi.org/10.18653/v1/K18-2008} {An improved neural
  network model for joint {POS} tagging and dependency parsing}.
\newblock In \emph{Proceedings of the {C}o{NLL} 2018 Shared Task: Multilingual
  Parsing from Raw Text to Universal Dependencies}, pages 81--91, Brussels,
  Belgium. Association for Computational Linguistics.

\bibitem[{Nivre et~al.(2016)Nivre, De~Marneffe, Ginter, Goldberg, Hajic,
  Manning, McDonald, Petrov, Pyysalo, Silveira et~al.}]{nivre2016universal}
Joakim Nivre, Marie-Catherine De~Marneffe, Filip Ginter, Yoav Goldberg, Jan
  Hajic, Christopher~D Manning, Ryan McDonald, Slav Petrov, Sampo Pyysalo,
  Natalia Silveira, et~al. 2016.
\newblock Universal dependencies v1: A multilingual treebank collection.
\newblock In \emph{Proceedings of the Tenth International Conference on
  Language Resources and Evaluation (LREC'16)}, pages 1659--1666.

\bibitem[{Peters et~al.(2018)Peters, Neumann, Iyyer, Gardner, Clark, Lee, and
  Zettlemoyer}]{peters-etal-2018-deep}
Matthew Peters, Mark Neumann, Mohit Iyyer, Matt Gardner, Christopher Clark,
  Kenton Lee, and Luke Zettlemoyer. 2018.
\newblock \href {https://doi.org/10.18653/v1/N18-1202} {Deep contextualized
  word representations}.
\newblock In \emph{Proceedings of the 2018 Conference of the North {A}merican
  Chapter of the Association for Computational Linguistics: Human Language
  Technologies, Volume 1 (Long Papers)}, pages 2227--2237, New Orleans,
  Louisiana. Association for Computational Linguistics.

\bibitem[{Rotman and Reichart(2019)}]{rotman2019deep}
Guy Rotman and Roi Reichart. 2019.
\newblock Deep contextualized self-training for low resource dependency
  parsing.
\newblock \emph{Transactions of the Association for Computational Linguistics},
  7:695--713.

\bibitem[{Sato et~al.(2017)Sato, Manabe, Noji, and
  Matsumoto}]{sato-etal-2017-adversarial}
Motoki Sato, Hitoshi Manabe, Hiroshi Noji, and Yuji Matsumoto. 2017.
\newblock \href {https://doi.org/10.18653/v1/K17-3007} {Adversarial training
  for cross-domain universal dependency parsing}.
\newblock In \emph{Proceedings of the {C}o{NLL} 2017 Shared Task: Multilingual
  Parsing from Raw Text to Universal Dependencies}, pages 71--79, Vancouver,
  Canada. Association for Computational Linguistics.

\bibitem[{Seeker and {\c{C}}etino{\u{g}}lu(2015)}]{seeker-cetinoglu-2015-graph}
Wolfgang Seeker and {\"O}zlem {\c{C}}etino{\u{g}}lu. 2015.
\newblock \href {https://doi.org/10.1162/tacl_a_00144} {A graph-based lattice
  dependency parser for joint morphological segmentation and syntactic
  analysis}.
\newblock \emph{Transactions of the Association for Computational Linguistics},
  3:359--373.

\bibitem[{Sigursson(2003)}]{sigurdhsson2003case}
Halld{\'o}r~{\'A}rmann Sigursson. 2003.
\newblock Case: abstract vs. morphological.
\newblock \emph{New perspectives on case theory}, pages 223--268.

\bibitem[{Stickland and Murray(2019)}]{pmlr-v97-stickland19a}
Asa~Cooper Stickland and Iain Murray. 2019.
\newblock \href {http://proceedings.mlr.press/v97/stickland19a.html} {{BERT}
  and {PAL}s: Projected attention layers for efficient adaptation in multi-task
  learning}.
\newblock In \emph{Proceedings of the 36th International Conference on Machine
  Learning}, volume~97 of \emph{Proceedings of Machine Learning Research},
  pages 5986--5995, Long Beach, California, USA. PMLR.

\bibitem[{Vania et~al.(2018)Vania, Grivas, and
  Lopez}]{vania-etal-2018-character}
Clara Vania, Andreas Grivas, and Adam Lopez. 2018.
\newblock \href {https://doi.org/10.18653/v1/D18-1278} {What do character-level
  models learn about morphology? the case of dependency parsing}.
\newblock In \emph{Proceedings of the 2018 Conference on Empirical Methods in
  Natural Language Processing}, pages 2573--2583, Brussels, Belgium.
  Association for Computational Linguistics.

\bibitem[{Vania et~al.(2019)Vania, Kementchedjhieva, S{\o}gaard, and
  Lopez}]{vania-etal-2019-systematic}
Clara Vania, Yova Kementchedjhieva, Anders S{\o}gaard, and Adam Lopez. 2019.
\newblock \href {https://doi.org/10.18653/v1/D19-1102} {A systematic comparison
  of methods for low-resource dependency parsing on genuinely low-resource
  languages}.
\newblock In \emph{Proceedings of the 2019 Conference on Empirical Methods in
  Natural Language Processing and the 9th International Joint Conference on
  Natural Language Processing (EMNLP-IJCNLP)}, pages 1105--1116, Hong Kong,
  China. Association for Computational Linguistics.

\bibitem[{Wunderlich and Lak{\"a}mper(2001)}]{wunderlich2001interaction}
Dieter Wunderlich and Renate Lak{\"a}mper. 2001.
\newblock On the interaction of structural and semantic case.
\newblock \emph{Lingua}, 111(4-7):377--418.

\bibitem[{Yang and Zhang(2018)}]{yang2018ncrf}
Jie Yang and Yue Zhang. 2018.
\newblock \href {http://aclweb.org/anthology/P18-4013} {Ncrf++: An open-source
  neural sequence labeling toolkit}.
\newblock In \emph{Proceedings of the 56th Annual Meeting of the Association
  for Computational Linguistics}.

\bibitem[{Zeman et~al.(2018)Zeman, Haji{\v{c}}, Popel, Potthast, Straka,
  Ginter, Nivre, and Petrov}]{zeman-etal-2018-conll}
Daniel Zeman, Jan Haji{\v{c}}, Martin Popel, Martin Potthast, Milan Straka,
  Filip Ginter, Joakim Nivre, and Slav Petrov. 2018.
\newblock \href {https://doi.org/10.18653/v1/K18-2001} {{C}o{NLL} 2018 shared
  task: Multilingual parsing from raw text to universal dependencies}.
\newblock In \emph{Proceedings of the {C}o{NLL} 2018 Shared Task: Multilingual
  Parsing from Raw Text to Universal Dependencies}, pages 1--21, Brussels,
  Belgium. Association for Computational Linguistics.

\bibitem[{Zhang et~al.(2015)Zhang, Zhao, and LeCun}]{NIPS2015_5782}
Xiang Zhang, Junbo Zhao, and Yann LeCun. 2015.
\newblock \href
  {http://papers.nips.cc/paper/5782-character-level-convolutional-networks-for-text-classification.pdf}
  {Character-level convolutional networks for text classification}.
\newblock In C.~Cortes, N.~D. Lawrence, D.~D. Lee, M.~Sugiyama, and R.~Garnett,
  editors, \emph{Advances in Neural Information Processing Systems 28}, pages
  649--657. Curran Associates, Inc.

\end{thebibliography}
\bibliographystyle{acl_natbib}
\appendix
\section*{Supplemental Material}
\label{base}
\section{Baselines}
\label{supple_baselines}
\subsection{BiAFFINE Parser (BiAFF)}
\label{supple_BiaFF}
BiAFF~\cite{DBLP:conf/iclr/DozatM17} is a graph-based dependency parsing approach similar to \newcite{kiperwasser-goldberg-2016-simple}. It uses biaffine attention instead of using a traditional MLP-based attention mechanism. For input vector $\vec{h}$, the affine classifier is expressed as $W\vec{h}+b$, while the biaffine classifier is expressed as $W'(W\vec{h}+b)+b'$. The choice of biaffine classifier facilitates the key benefit of representing the prior probability of word $j$ to be head and the likelihood of word $i$ getting word $j$ as the head. In this system, during training, each modifier in the predicted tree has the highest-scoring word as the head. This predicted tree need not be valid. However, at test time, to generate a valid tree MST algorithm~\cite{edmonds1967optimum} is used on the arc scores.

\subsection{Deep Contextualized Self-training (DCST)}
\label{supple_DCST}
~\newcite{rotman2019deep} proposed a self-training method called Deep Contextualized Self-training (DCST).\footnote{\url{https://github.com/rotmanguy/DCST}} 
In this system, the base parser BiAFF~\cite{DBLP:conf/iclr/DozatM17} is trained on the labelled dataset. 
Then this trained base parser is applied to the unlabelled data to generate automatically labelled dependency trees. In the next step, these automatically-generated trees are transformed into one or more sequence tagging schemes. Finally, the ensembled parser is trained on manually labelled data by integrating base parser with learned representation models.  The gating mechanism proposed by~\newcite{sato-etal-2017-adversarial} is used to integrate different tagging schemes into the ensembled parser. This approach is in line with the representation models based on language modeling related tasks~\cite{peters-etal-2018-deep,devlin-etal-2019-bert}. In summary, DCST demonstrates a novel approach to transfer information learned on labelled data to unlabelled data using sequence tagging schemes such that it can be integrated into final ensembled parser via word embedding layers.

\section{Experiments on TranSeq Variants}
\label{supple_TranSeq}

In TranSeq variations, instead of pretraining with three auxiliary tasks, we use a hierarchical multi-task morphological tagger~\cite{ashim2020evaluate} trained on 50k training data from DCS~\cite{hellwig2010dcs}.
In TranSeq setting, we extract the first three layers from this tagger and augment them in baseline models and experiment with five model sub-variants.
 \begin{table}[!hbt]
  \begin{small}
\centering
\begin{tabular}{ccccc}
\cmidrule(r){1-5}
 &\multicolumn{2}{c}{BiAFF}
&
\multicolumn{2}{c}{DCST} \\\cmidrule(r){2-3}\cmidrule(l){4-5}
Model &UAS&LAS    & UAS &LAS      \\\cmidrule(r){2-3}\cmidrule(l){4-5}

Base    & 70.67          & 56.85          & 73.23          & 58.64          \\
Base$\star$ & 69.35    &52.79 &    72.31&    54.82 \\
\cmidrule(r){2-3}\cmidrule(l){4-5}
TranSeq-FE  & 66.54          & 55.46          & 71.65          & 60.10          \\
TranSeq-FEA & 69.50          & 58.48          & 73.48          & 61.52          \\
TranSeq-UF  & 70.60          & 59.74          & 73.55          & 62.39          \\
TranSeq-DL  & 71.40          & 60.58          & 74.52          & 62.73 \\
TranSeq-FT  & \textbf{71.46} & \textbf{60.58} & \textbf{74.58} & \textbf{62.70}\\
\cmidrule(r){2-3}\cmidrule(l){4-5}
Oracle MI&\textit{74.08}&    \textit{62.48}&    \textit{76.66}&    \textit{66.35}\\

\hline
\end{tabular}
    \caption{Ablation analysis for TranSeq variations. Oracle MI is not comparable. It can be considered as upper bound for TranSeq.}
     \label{table:san_results}
     \end{small}
\end{table}
To avoid catastrophic forgetting~\cite{MCCLOSKEY1989109,french1999catastrophic}, we gradually increase the capacity of adaptations for each variant. \textbf{TranSeq-FE:} Freeze the pre-trained layers and use them as Feature Extractors (FE).  \textbf{TranSeq-FEA:} In the feature extractor setting, we additionally integrate adaptor modules~\cite{pmlr-v97-houlsby19a,pmlr-v97-stickland19a} in between two consecutive pre-trained layers.
     \textbf{TranSeq-UF:} Gradually Unfreeze (UF) these three pre-trained layers in the top to down fashion~\cite{felbo-etal-2017-using,howard-ruder-2018-universal}. \textbf{TranSeq-DL:} In this setting, we use discriminative learning (DL) rate~\cite{howard-ruder-2018-universal} for pre-trained layers, i.e., decreasing the learning rate as we move from top-to-bottom layers. \textbf{TranSeq-FT:} We fine-tune (FT), pre-trained layers with default learning rate used by~\newcite{DBLP:conf/iclr/DozatM17}.
     
    In the TranSeq setting, as we move down across its sub-variants in Table~\ref{table:san_results},  performance improves gradually, and TranSeq-FT configuration shows the best performance with 1-2 points improvement over Base. The Base$\star$ has one additional LSTM layer compared to Base such that the number of parameters are same as that of TranSeq-FT variation. The performance of Base$\star$ decreases compared to Base but TranSeq-FT outperforms Base. This shows that transfer learning definitely helps to boost the performance.
\end{document}